\documentstyle[11pt, epsf, subfigure]{article}

\newcommand{\postscript}[2]
{\setlength{\epsfxsize}{#2\hsize}
 \centerline{\epsfbox{#1}}}

\newcommand {\singlespace} {\addtolength{\baselineskip}{0.2\baselineskip}}

\newcommand{\mylinea}{\rule{6.2in}{0.5mm} \newline \vspace{0.2cm}}
\newcommand{\mylineb}{\rule{6.2in}{0.5mm} \newline}

\newtheorem{theorem}{Theorem}[section]

\newtheorem{definition}{Definition}[section]

\begin{document}
\thispagestyle{empty}
\begin{center}
~~ \\
\vspace{2in}
{\LARGE\bf Multi-Channel Parallel Adaptation Theory \\ 
for Rule Discovery} \\ ~~ \\
{\large\bf Li Min Fu} \\ 
 Department of CISE \\
 University of Florida  \\ ~~ \\ 
\end{center}
\begin{tabbing}
Corresponding Author: \\
Li Min Fu \\
Department of Computer and Information Sciences, 301 CSE \\
P.O. Box 116120 \\
University of Florida \\
Gainesville, Florida 32611 \\ ~~ \\
Phone: (352)392-1485 \\
e-mail: fu@cise.ufl.edu
\end{tabbing}
\newpage
\singlespace
\begin{center}
{\LARGE\bf Multi-Channel Parallel Adaptation Theory \\ 
for Rule Discovery} \\ 
{\large\bf Li Min Fu} \\
\end{center}
\begin{abstract}
In this paper, we introduce a new machine learning theory
based on multi-channel parallel adaptation for rule discovery. 
This theory is distinguished from the familiar parallel-distributed
adaptation theory of neural networks in terms of channel-based
convergence to the target rules.
We show how to realize this theory in a learning system named CFRule. 
CFRule is a parallel weight-based model, but it
departs from traditional neural computing
in that its internal knowledge is comprehensible. Furthermore,
when the model converges upon training,
each channel converges to a target rule.
The model adaptation rule is derived by multi-level parallel weight
optimization based on gradient descent.
Since, however, gradient descent only guarantees local optimization, 
a multi-channel regression-based optimization strategy is developed
to effectively deal with this problem.
Formally, we prove that the CFRule model can explicitly and precisely
encode any given rule set. Also, we prove a property related to
asynchronous parallel convergence, which is a critical element
of the multi-channel parallel adaptation theory for rule learning.
Thanks to the quantizability nature of the CFRule model, rules
can be extracted completely and soundly via a threshold-based mechanism.
Finally, the practical application of the theory is demonstrated
in DNA promoter recognition and hepatitis prognosis prediction.
\end{abstract}


\section{Introduction}
Rules express
general knowledge about actions or conclusions in given circumstances
and also principles in given domains. 
In the if-then format, rules are an easy way to represent
cognitive processes in psychology
and a useful means to encode expert knowledge. 
In another perspective,
rules are important because they can help scientists understand
problems and engineers solve problems.
These observations would account for the fact that
rule learning or discovery has become a major topic
in both machine learning and data mining research.
The former discipline concerns the construction of computer programs
which learn knowledge or skill while the latter is about the discovery of
patterns or rules hidden in the data. 

The fundamental concepts of rule learning are discussed in [16].
Methods for learning sets of rules include
symbolic heuristic search [3, 5],
decision trees [17-18],
inductive logic programming [13], neural networks [2, 7, 20],
and genetic algorithms [10]. 
A methodology comparison can be found in our previous work [9].
Despite the differences in their computational frameworks,
these methods perform a certain kind of search in 
the rule space (i.e., the space of possible rules) 
in conjunction with some optimization criterion.
Complete search is difficult unless the domain is small,
and a computer scientist is not interested in
exhaustive search due to its exponential computational
complexity. 
It is clear that significant issues have limited the effectiveness
of all the approaches described. 
In particular, we should point out that all the algorithms
except exhaustive search guarantee only local but
not global optimization.
For example,
a sequential covering algorithm such as CN2 [5] performs a greedy 
search for a single rule
at each sequential stage without backtracking and could make a
suboptimal choice at any stage;
a simultaneous covering algorithm such as ID3 [18] learns the entire
set of rules simultaneously 
but it searches incompletely through the
hypothesis space because of attribute ordering;
a neural network algorithm which adopts gradient-descent search 
is prone to local minima.

In this paper, we introduce a new machine learning theory
based on multi-channel parallel adaptation that shows great promise
in learning the target rules from data by parallel global convergence. 
This theory is distinct from the familiar parallel-distributed
adaptation theory of neural networks in terms of channel-based
convergence to the target rules.
We describe a system named CFRule which implements this theory.
CFRule bases its computational characteristics on the 
certain factor (CF) model [4, 22] it adopts.
The CF model is a calculus of uncertainty mangement and has been used
to approximate standard probability theory [1] in artificial intelligence.
It has been found that certainty factors associated with
rules can be revised by a neural network [6, 12, 15].
Our research has further indicated that
the CF model used as the neuron activation function (for combining inputs)
can improve the neural-network performance [8].

The rest of the paper is organized as follows.
Section~\ref{sec:mcrlm} describes the multi-channel rule learning model. 
Section~\ref{sec:mrr} examines the formal properties of rule encoding. 
Section~\ref{sec:mac} derives the model parameter adaptation rule,
presents a novel optimization strategy
to deal with the local minimum problem due to gradient descent,
and proves a property related to asynchronous
parallel convergence, which is a critical element
of the main theory.
Section~\ref{sec:re} formulates a rule extraction algorithm.
Section~\ref{sec:app} demonstrates practical applications. 
Then we draw conclusions in the final section.

\section{The Multi-Channel Rule Learning Model}
\label{sec:mcrlm}
CFRule is a rule-learning system based on multi-level parameter
optimization.
The kernel of CFRule is a multi-channel rule learning model.
CFRule can be embodied as an artificial neural network, but the neural
network structure is not essential.
We start with formal definitions about the model.

\begin{definition}
\label{def:mcrl}
The multi-channel rule learning model $M$ is defined by $k$ ($k \geq 1$)
channels ($Ch$'s), an input vector ($M_{in}$),
and an output ($M_{out}$)  as follows:
\begin{equation}
\label{eq:mcrl}
M \equiv (Ch_{1}, Ch_{2}, ..., Ch_{k}, M_{in}, M_{out})
\end{equation}
where $-1 \leq M_{out} \leq 1$ and
\begin{equation}
M_{in} \equiv (x_{1}, x_{2}, ..., x_{d})
\end{equation}
such that $d$ is the input dimensionality and
$-1 \leq x_{i} \leq 1$ for all $i$.
\end{definition}
The model has only a single output because here we assume the 
problem is a single-class, multi-rule learning problem.
The framework can be easily extended to the multi-class case.

\begin{definition}
\label{def:channel}
Each channel ($Ch_{j}$) is defined by an output weight ($u_{j}$),
a set of input weights ($w_{ji}$'s), activation ($\phi_{j}$), 
and influence ($\psi_{j}$) as follows:
\begin{equation}
Ch_{j} \equiv (u_{j}, w_{j0}, w_{j1}, w_{j2}, ..., w_{jd}, \phi_{j}, \psi_{j})
\end{equation}
where $w_{j0}$ is the bias,
$0 \leq u_{j} \leq 1$, and $-1 \leq w_{ji} \leq 1$ for all $i$.
The input weight vector $(w_{j1}, ..., w_{jd})$  defines
the channel's pattern.
\end{definition}

\begin{definition}
\label{def:chact}
Each channel's activation is defined by
\begin{equation}
\phi_{j} = f_{\rm cf}(w_{j0}, w_{j1}x_{1}, w_{j2}x_{2}, ..., w_{jd}x_{d})
\end{equation}
where $f_{\rm cf}$ is the CF-combining function [4, 22], as defined
below.
\end{definition}

\begin{definition}
\label{def:cf}
The CF-combining function is given by
\begin{equation}
\label{eq:cf}
f_{\rm cf}(x_{1}, x_{2}, ..., y_{1}, y_{2}, ...) = 
f_{\rm cf}^{+}(x_{1}, x_{2}, ...)
+ f_{\rm cf}^{-}(y_{1}, y_{2}, ...)
\end{equation}
where
\begin{equation}
f_{\rm cf}^{+}(x_{1}, x_{2}, ...) = 1 - \prod_{i} (1 - x_{i}) 
\end{equation}
\begin{equation}
f_{\rm cf}^{-}(y_{1}, y_{2}, ...) = -1 + \prod_{j} (1 + y_{j})
\end{equation}
$x_{i}$'s are nonnegative numbers and $y_{j}$'s are negative numbers.
\end{definition}
As we will see,
the CF-combining function contributes to several
important computational properties instrumental to rule discovery.

\begin{definition}
\label{def:chinf}
Each channel's influence on the output is defined by
\begin{equation}
\psi_{j} = u_{j}\phi_{j}
\end{equation}
\end{definition}

\begin{definition}
\label{def:output}
The model output $M_{out}$ is defined by
\begin{equation}
M_{out} = f_{\rm cf}(\psi_{1}, \psi_{2}, ..., \psi_{k})
\end{equation}
\end{definition}

We call the class whose rules to be learned the {\em target class},
and define rules inferring (or explaining) that class to be
the {\em target rules}. 
For instance, if the disease diabetes is the target class,
then the diagnostic rules for diabetes would be the target rules.
Each target rule defines a condition under which the given class
can be inferred. Note that we do not consider rules which deny
the target class, though such rules can be defined by reversing
the class concept. 
The task of rule learning is to learn or discover a set of
target rules from given instances called training instances (data).
It is important that rules learned should be generally 
applicable to the entire domain, not just the training data.
How well the target rules learned from the training data can be applied
to unseen data determines the generalization performance.

Instances which belong to the target class are called
positive instances, else, called negative instances.
Ideally, a positive training instance should match at least one target
rule learned and vice versa,
whereas a negative training instance should match none.
So, if there is only a single target rule learned, then it must be matched
by all (or most) positive training instances. 
But if multiple target rules are learned, then each rule
is matched by some (rather than all) positive training instances.
Since the number of possible rule sets is far greater than
the number of possible rules, the problem of learning multiple rules
is naturally much more complex than that of learning single rules.

In the multi-channel rule learning theory,
the model learns to sort out instances so that instances
belonging to different rules flow through different channels,
and at the same time, channels are adapted to accommodate their 
pertinent instances and learn corresponding rules.
Notice that this is a mutual process and it cannot occur all at once.
In the beginning, the rules are not learned and the channels
are not properly shaped, both information flow and adaptation
are more or less random, but through self-adaptation,
the CFRule model will gradually converge to the correct rules,
each encoded by a channel. 
The essence of this paper is to prove this property.

In the model design, a legitimate question is
what the optimal number of channels is.
This is just like the question raised for a neural network
of how many hidden (internal computing) units should be used.
It is true that too many hidden units cause
data overfitting and make generalization worse [7].
Thus, a general principle is to use a minimal number of hidden units.
The same principle can be equally well applied to the CFRule model.
However, there is a difference. In ordinary neural networks,
the number of hidden units is determined by the sample size,
while in the CFRule model, the number of channels
should match the number of rules embedded in the data.
Since, however, we do not know how many rules are present in the data,
our strategy is to use a minimal number of channels that
admits convergence on the training data.

The model's behavior is characterized by three aspects: 
\begin{itemize}
\item Information processing: Compute the model output for
a given input vector.
\item Learning or training: Adjust channels' parameters
(output and input weights) so that the input vector is mapped
into the output for every instance in the training data.
\item Rule extraction: Extract rules from a trained model.
\end{itemize}
The first aspect has been described already. 

\section{Model Representation of Rules}
\label{sec:mrr}
The IF-THEN rule (i.e., If the premise, then the action)
is a major knowledge representation paradigm in artificial intelligence.
Here we make analysis of how such rules can be represented with
proper semantics in the CFRule model.

\begin{definition}
\label{def:rule}
CFRule learns rules in the form of
\begin{center}
IF $A^{+}_{1}$, ..., $A^{+}_{i},$, ..., $\neg A^{-}_{1}$, . . ., 
$\neg A^{-}_{j}$, . . ., THEN the target class with a certainty factor. 
\end{center}
where $A^{+}_{i}$ is a positive antecedent (in 
the positive form), $A^{-}_{j}$ a negated antecedent
(in the negative form), and $\neg$ reads ``not.''
Each antecedent can be a discrete or discretized attribute (feature), 
variable, or a logic proposition. 
The IF part must not be empty.
The attached certainty factor in the THEN part, called the rule CF,
is a positive real $\leq 1$.
\end{definition}
The rule's premise is restricted to a conjunction,
and no disjunction is allowed.
The collection of rules for a certain class
can be formulated as a DNF (disjunctive normal form) logic expression, namely,
the disjunction of conjunctions, which implies the class.
However, rules defined here are not traditional logic rules
because of the attached rule CFs meant to capture uncertainty.
We interpret a rule by saying when its premise holds
(that is, all positive antecedents mentioned are true and
all negated antecedents mentioned are false), the target concept
holds at the given confidence level.
CFRule can also learn rules with weighted antecedents (a kind of
fuzzy rules), but we will not consider this case here.

There is increasing evidence to indicate
that good rule encoding capability actually
facilitates rule discovery in the data.
In the theorems that follow, we show how
the CFRule model can explicitly and precisely encode any given rule set.
We note that the ordinary sigmoid-function neural network can only
implicitly and approximately does this.
Also, we note although
the threshold function of the perceptron model enables it to learn
conjunctions or disjunctions,
the non-differentiability of this function prohibits
the use of an adaptive procedure in a multilayer construct.

\begin{theorem}
\label{thm:rule-rep}
For any rule represented by Definition~\ref{def:rule},
there exists a channel in the CFRule model to encode the rule so that
if an instance matches the rule, the channel's activation is 1, else 0. 
\end{theorem}
(Proof):
This can be proven by construction.
Suppose we implement channel $j$ by setting the bias weight to 1,
the input weights associated with
all positive attributes in the rule's premise to 1, 
the input weights associated with
all negated attributes in the rule's premise to $-1$, 
the rest of the input weights to 0,
and finally the output weight to the rule CF.
Assume that each instance is encoded by a bipolar vector in which
for each attribute, 1 means true and $-1$ false.
When an instance matches the rule, the following conditions hold:
$x_{i} = 1$ if $x_{i}$ is part of the rule's premise,
$x_{i} = -1$ if $\neg x_{i}$ is part of the rule's premise,
and otherwise $x_{i}$ can be of any value.
For such an instance, given the above construction, it is true that
$w_{ji}x_{i} = $ 1 or 0 for all $i$. 
Thus, the channel's activation (by Definition~\ref{def:chact}), 
\begin{equation}
\phi_{j} = f_{\rm cf}(w_{j0}=1, w_{j1}x_{1}, w_{j2}x_{2}, ..., w_{jd}x_{d})
\end{equation}
must be 1 according to $f_{cf}$.
On the other hand, if an instance does not match the rule, then there
exists $i$ such that $w_{ji}x_{i} = -1$.
Since $w_{j0}$ (the bias weight) = 1, the channel's
activation is 0 due to $f_{cf}$. $\Box$

\begin{theorem}
\label{thm:ruleset-rep}
Assume that rule CF's $> \theta$ ($0 \leq \theta \leq 1$).
For any set of rules represented by Definition~\ref{def:rule},
there exists a CFRule model to encode the rule set so that
if an instance matches any of the given rules, the model output is $> \theta$,
else 0. 
\end{theorem}
(Proof): Suppose there are $k$ rules in the set. 
As suggested in the proof of Theorem~\ref{thm:rule-rep}, 
we construct $k$ channels,
each encoding a different rule in the given rule set so that
if an instance matches, say rule $j$, then the activation ($\phi_{j}$) of
channel $j$ is 1. In this case,
since the channel's influence $\psi_{j}$
is given by $u_{j}\phi_{j}$ (where $u_{j}$
is set to the rule CF) and the rule CF $> \theta$, it follows that
$\psi_{j} > \theta$.
It is then clear that
the model output must be $> \theta$ since it
combines influences from all channels that $\geq 0$ but at least
one $> \theta$. On the other hand, if an instance fails to match any
of the rules, all the channels' activations are zero, so is
the model output.  $\Box$

\section{Model Adaptation and Convergence}
\label{sec:mac}
In neural computing, the backpropagation algorithm [19] can be viewed as
a multilayer, parallel optimization strategy that enables the
network to converge to a local optimum solution.
The black-box nature of the neural network solution is reflected
by the fact that the pattern (the input weight vector) learned by each
neuron does not bear meaningful knowledge. 
The CFRule model departs from traditional neural computing
in that its internal knowledge is comprehensible. Furthermore,
when the model converges upon training, 
each channel converges to a target rule. 
How to achieve this objective and what is the mathematical theory
are the main issues to be addressed.

\subsection{Model Training Based on Gradient Descent}
The CFRule model learns to map a set of input vectors
(e.g., extracted features) 
into a set of outputs (e.g., class information) by training. 
An input vector along with its target output constitute a training instance.
The input vector is encoded as a $1/-1$ bipolar vector.
The target output is 1 for a positive instance and 0 for
a negative instance.

Starting with a random or estimated weight setting,
the model is trained to adapt itself to the characteristics
of the training instances by changing weights
(both output and input weights) for every channel in the model.
Typically, instances are presented to the model one at a time.
When all instances are examined (called an epoch),
the network will start over with the first instance and repeat.
Iterations continue until the system performance 
has reached a satisfactory level.

The learning rule of
the CFRule model is derived in the same way as
the backpropagation algorithm [19].
The training objective is to minimize the sum of squared errors
in the data.
In each learning cycle, a training instance is given
and the weights of channel $j$ (for all $j$) are updated by
\begin{equation}
u_{j}(t+1) = u_{j}(t) + \Delta u_{j}
\end{equation}
\begin{equation}
w_{ji}(t+1) = w_{ji}(t) + \Delta w_{ji}
\end{equation}
where $u_{j}$: the output weight, $w_{ji}$: an input weight,
the argument $t$ denotes iteration $t$, and $\Delta$ the adjustment.
The weight adjustment on the current instance is based on gradient descent.
Consider channel $j$.
For the output weight ($u_{j}$),
\begin{equation}
\label{eq:gr-u}
\Delta u_{j} = - \eta (\partial E/ \partial u_{j})
\end{equation}
($\eta$: the learning rate) where
\[
E = \frac{1}{2}(T_{out} - M_{out})^{2}
\]
($T_{out}$: the target output, $M_{out}$: the model output).
Let 
\[
D = T_{out} - M_{out}
\]
The partial derivative in Eq.~(\ref{eq:gr-u})
can be rewritten with the calculus chain rule to yield
\[
\partial E/ \partial u_{j} = 
(\partial E/ \partial M_{out})(\partial M_{out}/ \partial u_{j}) =
-D (\partial M_{out}/ \partial u_{j})
\]
Then we apply this result to Eq.~(\ref{eq:gr-u}) and obtain
the following definition.

\begin{definition}
The learning rule for output weight $u_{j}$ of channel $j$ is given by
\begin{equation}
\label{eq:learn-u}
\Delta u_{j} =  \eta D (\partial M_{out}/ \partial u_{j})
\end{equation}
\end{definition}

For the input weights ($w_{ji}$'s), again based on gradient descent,
\begin{equation}
\label{eq:gr-w}
\Delta w_{ji} = - \eta (\partial E/ \partial w_{ji})
\end{equation}
The partial derivative in Eq.~(\ref{eq:gr-w}) is equivalent to
\[
\partial E/ \partial w_{ji} = 
(\partial E/ \partial \phi_{j})(\partial \phi_{j}/ \partial w_{ji})
\]
Since $\phi_{j}$ is not directly related to $E$, the first partial
derivative on the right hand side of the above equation is expanded
by the chain rule again to obtain
\[
\partial E/ \partial \phi_{j} =
(\partial E/ \partial M_{out})(\partial M_{out}/ \partial \phi_{j}) =
-D (\partial M_{out}/ \partial \phi_{j})
\]
Substituting these results into Eq.~(\ref{eq:gr-w})
leads to the following definition.

\begin{definition}
The learning rule for input weight $w_{ji}$ of channel $j$ is given by
\begin{equation}
\label{eq:learn-w}
\Delta w_{ji} =  \eta d_{j} (\partial \phi_{j} / \partial w_{ji})
\end{equation}
where
\[
d_{j} = D (\partial M_{out}/ \partial \phi_{j})
\]
\end{definition}

Assume that 
\begin{equation}
\phi_{j} = 
f_{\rm cf}^{+}(w_{j1}x_{1}, w_{j2}x_{2}, ..., w_{jd'}x_{d'})
+ f_{\rm cf}^{-}(w_{jd'+1}x_{d'+1}, ..., w_{jd}x_{d})
\end{equation}
Suppose $d' > 1$ and $d - d' > 1$.
The partial derivative $\frac{\partial \phi_{j}}{\partial w_{ji}}$
can be computed as follows. 
\begin{itemize}
\item[]Case (a)
If $w_{ji}x_{i} \geq 0$,
\begin{equation}
\frac{\partial \phi_{j}}{\partial w_{ji}} = (\prod_{l \neq i, l \leq d'}(1 - w_{jl}x_{l})) x_{i}
\label{eq:deriv1}
\end{equation}
\item[]Case (b)
If $w_{ji}x_{i} < 0$,
\begin{equation}
\frac{\partial \phi_{j}}{\partial w_{ji}} = (\prod_{l \neq i, l > d'}(1 + w_{jl}x_{l})) x_{i}
\label{eq:deriv2}
\end{equation}
\end{itemize}
It is easy to show that if $d' = 1$ in case (a) or $d - d' = 1$ in case (b),
$\frac{\partial \phi_{j}}{\partial w_{ji}} = x_{i}$.

\subsection{Multi-Channel Regression-Based Optimization}
\label{sec:mcro}
It is known that gradient descent can only find a local-minimum.
When the error surface is flat or very convoluted, 
such an algorithm often ends up with a bad local minimum.
Moreover, the learning performance is measured by the error on unseen data
independent of the training set. Such error is referred to
as generalization error.  We note that
minimization of the training error by the backpropagation algorithm
does not guarantee simultaneous minimization of generalization error.
What is worse, generalization error may instead rise after 
some point along the training curve due to an undesired phenomenon
known as overfitting [7].
Thus, global optimization techniques for network training (e.g.,
[21]) do not necessarily offer help as far as generalization is concerned.
To address this issue, CFRule uses a novel optimization
strategy called multi-channel regression-based optimization (MCRO).

In Definition~\ref{def:cf}, $f^{+}_{\rm cf}$ and $f^{-}_{\rm cf}$
can also be expressed as
\begin{equation}
f_{\rm cf}^{+}(x_{1}, x_{2}, ...) = 
\sum_{i} x_{i} - \sum_{i}\sum_{j} x_{i}x_{j} + \sum_{i}\sum_{j}\sum_{k}
x_{i}x_{j}x_{k} - ...
\end{equation}
\begin{equation}
f_{\rm cf}^{-}(y_{1}, y_{2}, ...) =
\sum_{i} y_{i} + \sum_{i}\sum_{j} y_{i}y_{j} + \sum_{i}\sum_{j}\sum_{k}
y_{i}y_{j}y_{k} + ...
\end{equation}
When the arguments ($x_{i}$'s and $y_{i}$'s) are small, the CF function
behaves somewhat like a linear function. It can be seen that if
the magnitude of every argument is $< 0.1$, the first order approximation
of the CF function is within an error of 10\% or so.
Since when learning starts, all the weights take on small values,
this analysis has motivated the MCRO strategy for improving
the gradient descent solution. The basic idea behind MCRO is
to choose a starting point based on the linear regression analysis,
in contrast to gradient descent which uses a random starting point.

If we can use regression analysis to estimate the initial 
influence of each input variable on the model output, how can we
know how to distribute this estimate over multiple channels? 
In fact, this is the most intricate part of the whole idea
since each channel's structure and parameters are yet to be learned.
The answer will soon be clear.

In CFRule, each channel's activation is defined by
\begin{equation}
\phi_{j} = f_{\rm cf}(w_{j0}, w_{j1}x_{1}, w_{j2}x_{2}, ...)
\end{equation}
Suppose we separate the linear component from the nonlinear component
($R$) in $\phi_{j}$ to obtain
\begin{equation}
\label{eq:phi-app}
\phi_{j} = (\sum_{i=0}^{d} w_{ji}x_{i}) + R_{j}
\end{equation}
We apply the same treatment to the model output (Definition~\ref{def:output})
\begin{equation}
M_{out} = f_{\rm cf}(u_{1}\phi_{1}, u_{2}\phi_{2}, ...)
\end{equation}
so that
\begin{equation}
\label{eq:sys-app}
M_{out} = (\sum_{j=1}^{k} u_{j}\phi_{j}) + R_{out}
\end{equation}
Then we substitute Eq.(\ref{eq:phi-app}) into Eq.(\ref{eq:sys-app})
to obtain
\begin{equation}
M_{out} = (\sum_{j=1}^{k}\sum_{i=0}^{d} u_{j}w_{ji}x_{i}) + R_{acc}
\end{equation}
in which the right hand side is equivalent to
\[
 [\sum_{i=0}^{d} (\sum_{j=1}^{k} u_{j}w_{ji})x_{i}] + R_{acc}
\]
Note that 
\[
R_{acc} = (\sum_{j=1}^{k} u_{j}R_{j}) + R_{out}
\]

Suppose linear regression analysis produces the following 
estimation equation for the model output:
\[
M_{out}' = b_{0} + b_{1}x_{1} + ...
\]
(all the input variables and the output transformed to the range from 0 to 1).
\begin{definition}
The MCRO strategy is defined by
\begin{equation}
\sum_{j=1}^{k} u_{j}(t=0)w_{ji}(t=0) = b_{i}
\end{equation}
for each $i, 0 \leq i \leq d$
\end{definition}
That is, at iteration $t=0$ when learning starts, the initial weights
are randomized but subject to these $d+1$ constraints.

\begin{table}
\begin{center}
\caption{The target rules in the simulation experiment.}
\label{tab:ex-rules}
\vspace{0.5cm}
\begin{tabular}{|lll|} \hline
rule 1: & IF $x_{1}$ and $\neg x_{2}$ and $x_{7}$ & THEN the target concept \\ 
rule 2: & IF $x_{1}$ and $\neg x_{4}$ and $x_{5}$ & THEN the target concept \\
rule 3: & IF $x_{6}$ and $x_{11}$ & THEN the target concept \\ \hline
\end{tabular}
\end{center}
\end{table}

To demonstrate this strategy, we designed an experiment.
Assume there were 20 input variables and 
three targets rules as shown in Table~\ref{tab:ex-rules}.
The training and test data sets were generated independently,
each consisting of 100 random instances.
An instance was classified as positive if it matched
any of the target rules and as negative otherwise.
The CFRule model for this experiment comprised three channels.
The model was trained under MCRO and random start separately.
For each strategy, 25 trials were run, each with
a different initial weight setting.
The same learning rate and stopping condition were used in
every trial regardless of the strategy taken. 
The training and test error rates were measured.
If the model converged to the target rules, then both training and
test errors should be close to zero.
We used the $t$ test (one-sided hypothesis testing based on
the statistical $t$ distribution) to evaluate the difference
in the means of error rates produced under the two strategies.
Given the statistical validation result (as summarized in
Table~\ref{tab:mcro}), we can conclude that MCRO is a valid technique.

\begin{table}
\begin{center}
\caption{Comparison of the MCRO strategy with random start 
for the convergence to the target rules. The results were validated
by the statistical $t$ test with the level of significance $< 0.01$
and $< 0.025$ (degrees of freedom = 48)
for the training and test error rates upon convergence, respectively.}
\label{tab:mcro}
\vspace{0.5cm}
\begin{tabular}{|l|c|c|c|c|} \hline
& {\em MCRO} & {\em Random Start} & {\em t-Value} & Level of Significance \\ \hline
Train error rate mean &   0.010 &    0.026 &  2.47  &   0.01 \\ 
Test error rate  mean &   0.012 &    0.033 &  2.34  &   0.025 \\ \hline
\end{tabular}
\end{center}
\end{table}

\subsection{Asynchronous Parallel Convergence}
\label{sec:converge}
In the multi-channel rule learning theory,
there are two possible modes of parallel convergence.
In the synchronous mode, all channels converge to
their respective
target patterns at the same time, whereas in the asynchronous
mode, each channel converges at a different time.
In a self-adaptation or self-organization model without a
global clock,
the synchronous mode is not a plausible scenario of convergence.
On the other hand, the asynchronous mode may not arrive at
global convergence (i.e., every channel converging to its target pattern) 
unless there is a mechanism to protect
a target pattern once it is converged upon.
Here we examine a formal property of CFRule on this new learning issue.

\begin{theorem}
\label{thm:converge}
Suppose at time $t$, channel $j$ of the CFRule model
has learned an exact pattern
$(w_{j1}, w_{j2}, ..., w_{jd})$ ($d \geq 1$)
such that $w_{j0}$ (the bias) $= 1$ 
and $w_{ji} = 1$ or $-1$ or 0 for $1 \leq i \leq d$.
At time $t+1$ when the model is trained on a given instance
with the input vector $(x_{0}, x_{1}, x_{2}, ..., x_{d})$ 
($x_{0} = 1$ and $x_{i} = 1$ or $-1$ for all $1 \leq i \leq d$),
the pattern is unchanged unless there is a single mismatched weight
(weight $w_{ji}$ is mismatched if and only if $w_{ji}x_{i} = -1$).
Let $\Delta w_{ji}(t+1)$ be the weight adjustment for $w_{ji}$.
Then \\
(a) If there is no mismatch, then $\Delta w_{ji}(t+1) = 0$ for all $i$. \\
(b) If there are more than one mismatched weight then 
$\Delta w_{ji}(t+1) = 0$ for all $i$.
\end{theorem}
(Proof): 
In case (a), there is no mismatch, so $w_{ji}x_{i} = 1$ or 0 for all $i$.
There exists $l$ such that $w_{jl}x_{l} = 1$ and $l \neq i$, for example,
$w_{j0}x_{0} = 1$ as given.
From Eq.~(\ref{eq:deriv1}),
\[
\frac{\partial \phi_{j}}{\partial w_{ji}} = 
(\prod_{l \geq 0, l \neq i}^{d}(1 - w_{jl}x_{l})) x_{i} = 0
\]
Then from Eq.~(\ref{eq:learn-w}),
\[
\Delta w_{ji}(t+1) = \eta d_{j}(\frac{\partial \phi_{j}}{\partial w_{ji}}) = 0
\]
In case (b), the proof for matched weights is the same as that in case (a).
Consider only mismatched weights $w_{ji}$'s such that $w_{ji}x_{i} = -1$.
Since there are at least two mismatched weights, 
there exists $l$ such that $w_{jl}x_{l} = -1$ and $l \neq i$.
From Eq.~(\ref{eq:deriv2}),
\[
\frac{\partial \phi_{j}}{\partial w_{ji}} = 
(\prod_{w_{jl}x_{l} = -1, l \neq i}(1 + w_{jl}x_{l})) x_{i} = 0
\]
Therefore,
\[
\Delta w_{ji}(t+1) = \eta d_{j}(\frac{\partial \phi_{j}}{\partial w_{ji}}) = 0
\]
In the case of a single mismatched weight,
\[
\frac{\partial \phi_{j}}{\partial w_{ji}} = x_{i}
\]
which is not zero, so the weight adjustment $\Delta w_{ji}(t+1)$
may or may not be zero, depending on the error $d_{j}$. $\Box$

Since model training starts with small weight values, the initial
pattern associated with each channel cannot be exact. When training
ends, the channel's pattern may still be inexact because of
possible noise, inconsistency, and uncertainty in the data.
However, from the proof of the above theorem, we see that when
the nonzero weights in the channel's pattern grow larger, the error
derivative ($d_{j}\frac{\phi_{j}}{w_{ji}}$) generally gets smaller,
so does the weight adjustment, and as a result, the pattern becomes
more stable and gradually converges to a target pattern.
A converged pattern does not move
unless there is a near-miss instance 
(with a single feature mismatch against the pattern)
that causes some error in the model
output, in which case, the pattern is refined to be a little more
general or specific.
This analysis explains how the CFRule model
ensures the stability of a channel once it is settled in 
a target pattern.
Note that the output weight of a channel
with a stable pattern can still be modified toward global
error minimization and uncertainty management.
In asynchronous parallel convergence, each channel is settled in
its own target pattern with a different time frame. 
Without the above pattern stabilizing property, global convergence 
is difficult to achieve in the asynchronous mode. 
This line of arguments imply that CFRule admits asynchronous
parallel convergence.
Theorem~\ref{thm:converge} is unique for CFRule. 
That property has not been provable for other types of neural networks
or learning methods (e.g., [16]).

\begin{figure}
\postscript{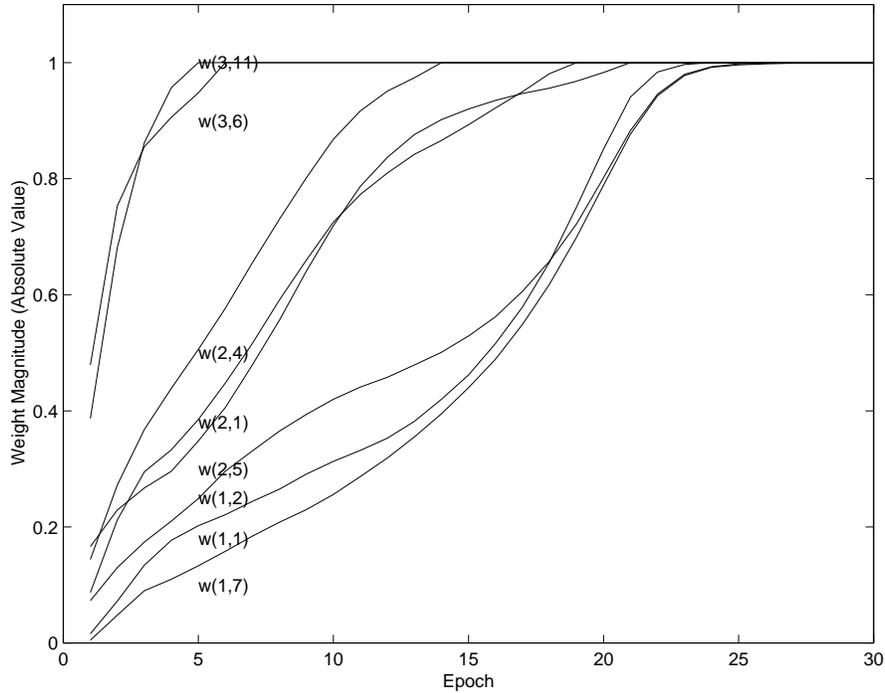}{0.75}
\caption{The temporal curves of asynchronous parallel convergence
for rule learning.}
\label{fig:converge}
\end{figure}

Asynchronous parallel convergence for rule learning can be
illustrated by the example in Section~\ref{sec:mcro}.
Table~\ref{tab:converge} shows how each channel converges to
a target rule in the training course
when the model was trained on just 100 random instances (out of
$2^{20}$ possible instances).
For instance, given $\neg x_{2}$ in the premise of rule 1
(Table~\ref{tab:ex-rules}), we observe 
the corresponding weight $w_{1,2}$ of channel 1 converged to $-1$ 
(Table~\ref{tab:converge});
also, for $x_{6}$ mentioned in rule 3, we see
the weight $w_{3,6}$ of channel 3 converged to 1.
Only the significant weights that converge to a magnitude of
1 are shown. Unimportant weights ending up with about zero values
are omitted.
The convergence behavior can be better visualized in
Figure~\ref{fig:converge}. It clearly shows that convergence
occurs asynchronously for each channel. 
It does not matter which channel converges to which rule.
This correspondence is determined by the initial weight setting
and the data characteristics. Note that
given $k$ channels in the model,
there are $k!$ equivalent permutations in terms of their relative
positions in the model. 
It matters, though, whether the model as a whole
converges to all the needed target rules.

\begin{table}
\begin{center}
\caption{Asynchronous parallel convergence to the target rules
in the CFRule model. Channels 1, 2, 3 converge to
target rules 1, 2, 3, respectively. $w_{j,i}$ denotes the
input weight associated with the input $x_{i}$ in channel $j$.
An epoch consists of a presentation of all training instances.}
\label{tab:converge}
\vspace{0.5cm}
\begin{tabular}{|c|cccccccc|} \hline
{\em epoch} & $w_{1,1}$ & $w_{1,2}$ & $w_{1,7}$ & $w_{2,1}$ & $w_{2,4}$ 
& $w_{2,5}$ & $w_{3,6}$ & $w_{3,11}$  \\ \hline
1 & .016 & -.073 & .005 & .144 & -.166 & .087 & .387 & .479 \\ \hline
5 & .202 & -.249 & .133 & .506 & -.348 & .385 & 1.00 & .948 \\ \hline
10 & .313 & -.420 & .256 & .868 & -.719 & .725 & 1.00 & 1.00 \\ \hline
15 & .462 & -.529 & .440 & 1.00 & -.920 & .893 & 1.00 & 1.00 \\ \hline
20 & .851 & -.802 & .789 & 1.00 & -.983 & 1.00 & 1.00 & 1.00 \\ \hline
25 & 1.00 & -.998 & .996 & 1.00 & -1.00 & 1.00 & 1.00 & 1.00 \\ \hline
30 & 1.00 & -1.00 & 1.00 & 1.00 & -1.00 & 1.00 & 1.00 & 1.00 \\ \hline
\end{tabular}
\end{center}
\end{table}

\section{Rule Extraction}
\label{sec:re}
As illustrated by the example in Section~\ref{sec:converge},
when a channel converges
to a target rule, the weights associated with the input attributes
contained in the rule's premise grow into large values, whereas the rest
of input weights decay to small values. 
The asymptotic absolute weight values
upon convergence approach either 1 or 0 ideally,
but this case does not necessarily happen in practical circumstances
involving data noise, inconsistency, uncertainty, and inadequate
sample sizes.
However, in whatever circumstances, it turns out that 
a simple thresholding mechanism suffices to distinguish important
from unimportant weights in the CFRule model. 
Since the weight absolute values range from 0 to 1, it is
reasonable to use 0.5 as the threshold, but this value does not always
guarantee optimal performance. How to search for a good threshold
in a continuous range is difficult. Fortunately, thanks to
the quantizability nature of the system adopting the CF model [9],
only a handful of values need to be considered. Our research has narrowed
it down to four candidate values: 0.35, 0.5, 0.65, and 0.8.
A larger threshold makes extracted rules more general, whereas
a smaller threshold more specific.
In order to lessen data overfitting, our heuristic is to choose
a higher value as long as the training error is acceptable.
Using an independent cross-validation data set
is a good idea if enough data is available.
The rule extraction algorithm is formulated below.

\begin{center}
\mylinea
{\em The CFRule Rule Extraction Algorithm} 
\begin{itemize}
\item Select a rule extraction threshold $r$ ($0 < r <1$).
\item For each channel $j$, 
\begin{enumerate}
\item $P := nil$ (an empty set)
\item $C := $the target class
\item Normalize the input weights $w_{ji}$'s so that the maximum weight
absolute value is 1. 
\item For each input weight $w_{ji}$
($1 \leq i \leq d$, $d$: the input dimensionality), 
\begin{itemize}
\item[a.]
If $w_{ji} \geq r$, then add $x_{i}$ to $P$.
\item[b.]
If $w_{ji} \leq -r$, then add $\neg x_{i}$ to $P$.
\item[c.]
Else, do nothing.
\end{itemize}
\item Form a rule: ``IF $P$, THEN $C$ with CF = $u_{j}$'' 
($u_{j}$: the output weight based on the rule).
\end{enumerate}
\item Remove subsumed rules and rules with low CFs.
\end{itemize}
\mylineb
\end{center}

The threshold-based algorithm described here
is fundamentally different from the search-based
algorithm in neural network rule extraction [7, 9, 20].
The main advantage with the threshold-based approach is
its linear computational complexity 
with the total number of weights,
in contrast to polynomial
or even exponential complexity incurred by the search-based approach.
Furthermore, the former approach obviates the need of a special
training, pruning, or approximation procedure commonly used in
the latter approach for complexity reduction. 
As a result, the threshold-based, direct approach should
produce better and more reliable rules.
Notice that this approach is not applicable
to the ordinary sigmoid-function neural network where knowledge
is entangled. 
The admissibility of the threshold-based algorithm for rule
extraction in CFRule can be ascribed to the CF-combining function.

\section{Applications}
\label{sec:app}
Two benchmark data sets were selected to demonstrate the value of
CFRule on practical domains.
The promoter data set is characterized by high dimensionality
relative to the sample size, while the hepatitis data has a lot of
missing values. Thus, both pose a challenging problem.

The decision-tree-based rule generator system C4.5 [18] was
taken as a control since it (and with its later version)
is the currently most representative (or most often used)
rule learning system, and also 
the performance of C4.5 is optimized in a statistical sense.

\subsection{Promoter Recognition in DNA}
In the promoter data set [23], there are 106 instances with
each consisting of a DNA nucleotide string of four
base types: A (adenine), G (guanine), 
C (cytosine), and T (thymine).
Each instance string is comprised of 57 sequential nucleotides,
including fifty nucleotides before (minus) and six following (plus)
the transcription site. 
An instance is a positive instance if the promoter region
is present in the sequence, else it is a negative instance.
There are 53 positive instances and 53 negative instances, respectively. 
Each position of an instance sequence is
encoded by four bits with each bit designating a base type.
So an instance is encoded by a vector of 228 bits
along with a label indicating a positive or negative instance.

In the literature of molecular biology,
promoter (of prokaryotes)
sequences have average constitutions of -TTGACA-
and -TATAAT-, respectively, located at so-called minus-35 and
minus-10 regions [14], as shown in Table~\ref{tab:prom-domain-rule}.

\begin{table}
\begin{center}
\caption{The promoter (of prokaryotes) consensus sequences.}
\label{tab:prom-domain-rule}
 \vspace{0.5cm}
\begin{tabular}{|c|l|} \hline
{\em Region} & {\em DNA Sequence Pattern}
\\ \hline \hline
Minus-35 & 
@-36=T @-35=T @-34=G @-33=A \\
& @-32=C @-31=A  \\ \hline
Minus-10 &
@-13=T @-12=A @-11=T @-10=A \\
& @-9=A  @-8=T  \\ \hline
\end{tabular}
\end{center}
\end{table}

The CFRule model in this study had 3 channels,
which were the minimal number of channels to bring
the training error under 0.02 upon convergence.
Still, the model 
is relatively underdetermined because of the low ratio of the 
number of instances available for training to the input dimension. 
However, unlike our previous approach [9], we did not use any pruning
strategy. The model had to learn to cope with high dimensionality
by itself. The learning rate was set to 0.2, and the rule extraction
threshold 0.5 (all these are default values). The model was trained
on the training data under the MCRO strategy and then tested on the
test data. The stopping criterion for training was the
drop of MSE (mean squared error) less than a small value per epoch.
Rules were extracted from the trained model.

Cross-validation is an important means to evaluate the
ability of learning.
Domain validity is indicated if rules learned based on some data
can be well applied to other data in the same domain.
In the two-fold cross-validation experiment, the 106 instances
were randomly divided equally into two subsets.
CFRule and C4.5 used the same data partition.
The rules learned on one subset were tested by the other
and vice versa.
The average prediction error rate on the test set was defined as
the cross-validation {\em rule} error rate.
The cross-validation experiment with CFRule
was run 5 times, each with a different
initial weight setting. The average cross-validation error was reported.
CFRule had a significantly smaller cross-validation rule
error rate than C4.5 (12.8\% versus 23.9\%, respectively),
as shown in Table~\ref{tab:cv}.
Note that the prediction accuracy and the error rate were measured
based on exact symbolic match. That is, an instance is predicted
to be in the concept only if it matches exactly any rule of the concept,
else it is not in the concept.
If, however, prior domain knowledge is used 
and exact symbolic match is not required, 
the error rate based on leave-one-out
can be as low as 2\% [7].

\begin{table}
\begin{center}
\caption{The average two-fold cross-validation error rates of the rules
learned by C4.5 and CFRule, respectively.}
\label{tab:cv}
\vspace{0.5cm}
\begin{tabular}{|l|c|c|} \hline
{\em Domain} & {\em C4.5} & {\em CFRule} \\ \hline \hline
Promoters (without & & \\
prior knowledge) & 23.9\% & 12.8\% \\ \hline
Hepatitis & 7.1\% & 5.3\% \\ \hline
\end{tabular}
\end{center}
\end{table}

Both CFRule and C4.5 learned three rules from the 106 instances.
The rules are summarized in Table~\ref{tab:prom-rule}. 
In the aspect of rule quality, 
CFRule learned rules of larger size than C4.5 under inadequate
samples. This is because CFRule tends to keep attributes sufficiently
correlated with the target concept, whereas C4.5 retains only 
attributes with verified statistical significance and tends 
to favor more general rules.
In terms of domain validity,
the rule's accuracy based on cross-validation
is more reliable than other quality measures.
Another interesting discovery made by CFRule (but not by C4.5) is
@-45=A (in rule \#3) which plays a major role in the so-called
conformation theory for promoter prediction [11].

\begin{table}
\begin{center}
\caption{The promoter prediction rules learned from 106 instances
by CFRule and C4.5, respectively.  $\neg$: not. @: at.}
\label{tab:prom-rule}
 \vspace{0.5cm}
\begin{tabular}{|l|c|l|} \hline
& {\em Rule} & {\em DNA Sequence Pattern}
\\ \hline \hline
CFRule & \# 1 &
@-34=G @-33=$\neg$G @-12=$\neg$G  \\ 
& \# 2 & 
@-36=T @-35=T @-31=$\neg$C @-12=$\neg$G \\
& \# 3 &
@-45=A @-36=T @-35=T  \\ \hline
C4.5 & \# 1 & 
@-35=T @-34=G \\  
& \# 2 &
@-36=T @-12=A \\ 
& \# 3 & 
@-36=T @-35=T @-34=T \\  \hline
\end{tabular}
\end{center}
\end{table}

The data for this research are
available from a machine learning database located in the
University of California at Irvine with an ftp address at
ftp.ics.uci.edu/pub/machine-learning-databases.

\subsection{Hepatitis Prognosis Prediction}
In the data set concerning hepatitis prognosis 
\footnote{This data set is an old version previously used in our
research work [7].},
there are 155 instances, each described by 19 attributes.
Continuous attributes were discretized, then
the data set was randomly partitioned into two
halves (78 and 77 cases), and then cross-validation was carried out.
CFRule and C4.5 used exactly the same data to ensure fair comparison.
The CFRule model for this problem consisted of 2 channels.
Again, CFRule was superior to C4.5 based on the cross-validation performance
(see Table~\ref{tab:cv}).
However, both systems learned the same single rule
from the whole 155 instances, as displayed in Table~\ref{tab:hepa-rule}.
To learn the same rule by two fundamentally different systems
is quite a coincidence, but it suggests the rule is true in a global sense.

\begin{table}
\begin{center}
\caption{The hepatitis rules for predicting (bad) prognosis 
learned from 155 instances
by CFRule and C4.5, respectively.  $\neg$: not. @: at.}
\label{tab:hepa-rule}
 \vspace{0.5cm}
\begin{tabular}{|l|c|l|} \hline
& {\em Rule} & {\em Premise}
\\ \hline \hline
CFRule 
& \# 1 & MALE and NO STEROID and ALBUMIN $<$ 3.7 \\ \hline
C4.5  
& \# 1 & MALE and NO STEROID and ALBUMIN $<$ 3.7 \\ \hline
\end{tabular}
\end{center}
\end{table}

\section{Conclusions}
If global optimization is a main issue for automated rule discovery
from data, then current machine learning theories do not seem adequate.
For instance, the decision-tree and neural-network based algorithms,
which dodge the complexity of exhaustive search,
guarantee only local but not global optimization. 
In this paper, we introduce a new machine learning theory
based on multi-channel parallel adaptation that shows great promise
in learning the target rules from data by parallel global convergence. 
The basic idea is that when
a model consisting of multiple
parallel channels is optimized according to a certain global
error criterion, each of its channels converges to a target rule.
While the theory sounds attractive, the main question is how to implement it.
In this paper, we show how to realize this theory in a learning system
named CFRule.

CFRule is a parallel weight-based model, which can be optimized
by weight adaptation. 
The parameter adaptation rule follows the gradient-descent idea
which is generalized
in a multi-level parallel context. However, the central idea of
the multi-channel rule-learning theory is not about how the parameters
are adapted but rather, how each channel can converge to
a target rule. We have noticed that CFRule exhibits the necessary
conditions to ensure such convergence behavior. 
We have further found that the CFRule's behavior can be attributed to
the use of the CF (certainty factor) model for combining
the inputs and the channels. 

Since the gradient descent technique seeks only a local minimum,
the learning model may well be settled in a solution where each
rule is optimal in a local sense. 
A strategy called multi-channel regression-based optimization (MCRO)
has been developed to address this issue. 
This strategy has proven effective by statistical validation.

We have formally proven two important properties that account for the
parallel rule-learning behavior of CFRule. 
First, we show that any given rule set can be explicitly and
precisely encoded by the CFRule model. Secondly, we show that
once a channel is settled in a target rule, it barely moves.
These two conditions encourage the model to move toward the
target rules. An empirical weight convergence graph clearly showed
how each channel converged to a target rule in an asynchronous manner.
Notice, however, we have not been able to prove or demonstrate
this rule-oriented convergence behavior in other neural networks.

We have then examined the application of this methodology
to DNA promoter recognition and hepatitis prognosis prediction.
In both domains, CFRule is superior to C4.5 
(a rule-learning method based on the decision tree)
based on cross-validation. Rules learned are also consistent with
knowledge in the literature.

In conclusion, the multi-channel parallel adaptive rule-learning 
theory is not just theoretically sound and supported by
computer simulation but also practically useful.
In light of its significance, 
this theory would hopefully point out a new direction for
machine learning and data mining.

\newpage
\begin{flushleft}
{\large\bf Acknowledgments}
\end{flushleft}
This work is supported by National Science
Foundation under the grant ECS-9615909.
The author is deeply indebted to Edward Shortliffe who contributed
his expertise and time in the discussion of this paper.

\begin{flushleft}
{\Large\bf References}
\end{flushleft}
\begin{enumerate}
\item 
J.B. Adams, ``Probabilistic reasoning and certainty factors'',
in {\em Rule-Based Expert Systems}, Addison-Wesley, Reading, MA, 1984.
\item 
J.A. Alexander and M.C. Mozer,
``Template-based algorithms for connectionist rule extraction'',
in {\em Advances in Neural Information Processing Systems},
MIT Press, Cambridge, MA, 1995.
\item 
 B.G. Buchanan and T.M. Mitchell,
 ``Model-directed learning of production 
rules'', in {\em Pattern-Directed Inference Systems}, 
Academic Press, New York, 1978.
\item
B.G. Buchanan and E.H. Shortliffe (eds.),
{\em Rule-Based Expert Systems}, Addison-Wesley, Reading, MA, 1984.
\item 
P. Clark and R. Niblett,
``The CN2 induction algorithm'',
{\em Machine Learning}, 3, pp. 261-284, 1989.
\item L.M. Fu,  ``Knowledge-based connectionism for revising domain
 theories'', 
{\em  IEEE Transactions on Systems, Man, and Cybernetics}, 23(1),
pp. 173{\rm --}182, 1993.
\item
L.M. Fu, 
{\em Neural Networks in Computer Intelligence}, 
McGraw Hill, Inc., New York, NY, 1994.
\item
L.M. Fu, 
``Learning in certainty factor based 
multilayer neural networks for classification'',
{\em IEEE Transactions on Neural Networks}. 9(1), pp. 151-158, 1998.
\item
L.M. Fu and E.H. Shortliffe,
``The application of certainty factors to neural computing
for rule discovery'',
{\em IEEE Transactions on Neural Networks},  11(3), pp. 647-657, 2000.
\item
C.Z. Janikow,
``A knowledge-intensive genetic algorithm for supervised learning'',
{\em Machine Learning}, 13, pp. 189-228, 1993.
\item
G.B. Koudelka, S.C. Harrison, and M. Ptashne, 
``Effect of non-contacted bases on the affinity of
434 operator for 434 repressor and Cro'',
{\em Nature}, 326, pp. 886-888, 1987.
\item
R.C. Lacher, S.I. Hruska, and D.C. Kuncicky,
``Back-propagation learning in expert networks'', 
{\em IEEE Transactions on Neural Networks}, 3(1), pp. 62{\rm --}72,
1992.
\item
N. Lavra\u{c} and S. D\u{z}eroski, 
{\em Inductive Logic Programming: Techniques and Applications},
Ellis Horwood, New York, 1994.
\item S.D. Lawrence, 
{\em The Gene}, Plenum Press, New York, NY, 1987.
\item
J.J. Mahoney and R. Mooney, 
``Combining connectionist and symbolic learning to refine
certainty-factor rule bases'', {\em Connection Science}, 5,
pp. 339-364, 1993. 
\item
T. Mitchell, {\em Machine learning}, 
McGraw Hill, Inc., New York, NY., 1997. 
\item
J.R. Quinlan, 
``Rule induction with statistical data---a comparison with
multiple regression'', 
{\em Journal of the Operational Research Society},
38, pp. 347-352, 1987.
\item
J.R. Quinlan, {\em C4.5: Programs for Machine Learning},
Morgan Kaufmann, San Mateo, CA., 1993.
\item 
D.E. Rumelhart,  G.E. Hinton,  and R.J. Williams,
``Learning internal representation by error propagation'',
In {\em Parallel Distributed Processing: Explorations in the
Microstructures of Cognition}, Vol. 1.
MIT press, Cambridge, MA, 1986.
\item
R. Setiono and H. Liu, ``Symbolic representation of neural networks'',
{\em Computer}, 29(3), pp. 71-77, 1996.
\item
Y. Shang and B.W. Wah, ``Global optimization for neural network training'',
{\em Computer}, 29(3), pp. 45-54, 1996.
\item
E.H. Shortliffe and B.G. Buchanan, ``A model of inexact reasoning in
medicine'', {\em Mathematical Biosciences}, 23, pp. 351-379, 1975.
\item
G.G. Towell and J.W. Shavlik,
``Knowledge-based artificial neural networks'', 
{\em Artificial Intelligence}. 70(1-2), pp. 119-165, 1994.
\end{enumerate}

\end{document}